\documentclass{article}

\usepackage{microtype}
\usepackage{graphicx}
\usepackage{subfigure}
\usepackage{booktabs} 
\usepackage{adjustbox}
\usepackage{makecell} 
\usepackage{pdflscape}  
\usepackage{longtable}  
\usepackage{rotating}
\usepackage{pgfplots}
\usetikzlibrary{pgfplots.groupplots}

\usepackage{hyperref}

\usepackage[accepted]{icml2024}

\usepackage{amsmath}
\usepackage{amssymb}
\usepackage{mathtools}
\usepackage{amsthm}

\usepackage[capitalize,noabbrev]{cleveref}

\theoremstyle{plain}

\theoremstyle{definition}

\theoremstyle{remark}

\usepackage[textsize=tiny]{todonotes}

\usepackage{titlesec}
\titlespacing{\paragraph}{%
  0pt}{
  0pt}{
  1em}

\usepackage{xspace}
\newcommand{\method}{{\fontfamily{lmtt}\selectfont PHATGOOSE}\xspace}

\definecolor{light-gray}{gray}{0.7}

\icmltitlerunning{Learning to Route Among Specialized Experts for Zero-Shot Generalization}

\begin{document}

\twocolumn[
\icmltitle{Learning to Route Among Specialized Experts for Zero-Shot Generalization}



\icmlsetsymbol{equal}{*}

\begin{icmlauthorlist}
\icmlauthor{Mohammed Muqeeth}{MIT-IBM}
\icmlauthor{Haokun Liu}{UofT,Vector}
\icmlauthor{Yufan Liu}{UNC}
\icmlauthor{Colin Raffel}{UofT,Vector}
\end{icmlauthorlist}

\icmlaffiliation{MIT-IBM}{MIT-IBM}
\icmlaffiliation{UNC}{University of North Carolina at Chapel Hill}
\icmlaffiliation{UofT}{University of Toronto}
\icmlaffiliation{Vector}{Vector Institute}

\icmlcorrespondingauthor{Mohammed Muqeeth}{muqeeth101@gmail.com}
\icmlcorrespondingauthor{Haokun Liu}{haokunliu412@gmail.com}
\icmlcorrespondingauthor{Colin Raffel}{craffel@gmail.com}

\icmlkeywords{Machine Learning, ICML}

\vskip 0.3in
]



\printAffiliationsAndNotice{}  

\begin{abstract}
    Recently, there has been a widespread proliferation of ``expert'' language models that are specialized to a specific task or domain through parameter-efficient fine-tuning.
    How can we recycle large collections of expert language models to improve zero-shot generalization to unseen tasks?
    In this work, we propose \textbf{P}ost-\textbf{H}oc \textbf{A}daptive \textbf{T}okenwise \textbf{G}ating \textbf{O}ver an \textbf{O}cean of \textbf{S}pecialized \textbf{E}xperts (\method), which learns to route among specialized modules that were produced through parameter-efficient fine-tuning.
    Unlike past methods that learn to route among specialized models, \method explores the possibility that zero-shot generalization will be improved if different experts can be adaptively chosen for each token and at each layer in the model.
    Crucially, our method is \textit{post-hoc} - it does not require simultaneous access to the datasets used to create the specialized models and only requires a modest amount of additional compute after each expert model is trained.
    In experiments covering a range of specialized model collections and zero-shot generalization benchmarks, we find that \method outperforms past methods for post-hoc routing and, in some cases, outperforms explicit multitask training (which requires simultaneous data access).
    To better understand the routing strategy learned by \method, we perform qualitative experiments to validate that \method's performance stems from its ability to perform per-token and per-module routing.
    We release all of our code to support future work on improving zero-shot generalization by recycling specialized experts.\footnote[1]{\label{note:code} \url{https://github.com/r-three/phatgoose}}
\end{abstract}

\section{Introduction}

The availability of performant pre-trained language models has led to a proliferation of fine-tuned ``expert'' models that are specialized to a particular task or domain.
Many of these expert models are created through parameter-efficient fine-tuning (PEFT) techniques \cite{ding2022delta,lialin2023scaling,he2021towards}, which produce a fine-tuned model by adding small ``modules'' (such as Low-Rank Adapters \citep{hu2021lora} or (IA)$^3$ vectors \citep{liu2022few}) that only introduce or modify a small number of parameters.
Specialized PEFT modules can be easily shared due to their small size, which has led to the distribution of an ever-growing number of adapters on various platforms -- for example, as of writing over 17,000 adapters based on the \texttt{peft} library \cite{peft} have been uploaded to the Hugging Face Model Hub.\footnote{\url{https://huggingface.co/models?library=peft}}
The availability of these PEFT modules makes it cheap and easy to modularly adapt a given pre-trained model to a specific task or domain.

\begin{figure}[t]
\centering
\includegraphics[width=0.9\columnwidth]{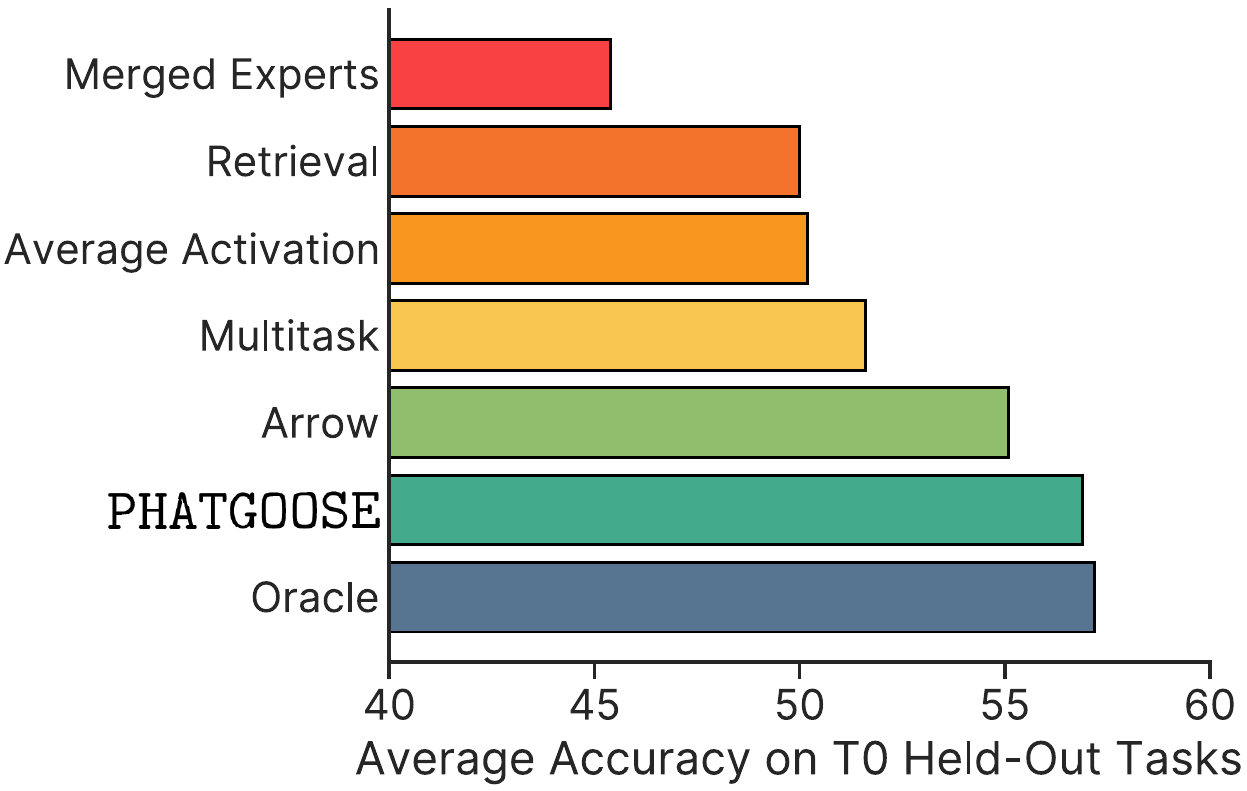}
\vspace{-0.5em}
\caption{Average performance of different multitask training and expert routing methods when using the same held-in and held-out tasks as T0 \citep{sanh2021multitask}. Notably, our proposed method \method outperforms all past methods for recycling experts as well as explicit multitask training (which requires simultaneous data access) and nearly matches the performance of an oracle routing scheme. See \cref{sec:experiments} for more details. Exact numerical results for all methods can be found in \cref{tab:T0_results_all_methods}.}
\vspace{-1em}
\label{fig:t0_results}
\end{figure}

\begin{figure*}[t]
\centering
\includegraphics[width=0.7\textwidth]{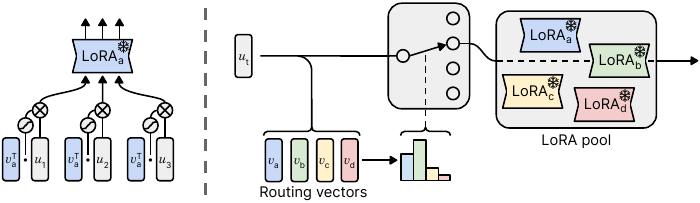}
\vspace{-0.5em}
\caption{Visualization of how \method learns to route among specialized modules. This diagram shows how routing is learned at a layer where a module is inserted; typically a PEFT-based model introduces many such modules at  various layers throughout the model. Left: After a specialized module (here, shown as a LoRA \citep{hu2021lora} module) has been trained, it is frozen and a sigmoid gate is trained to choose which activations should be fed into the module. Right: During inference, a routing distribution (shown as a bar plot) is computed from the dot product scores between the normalized gates and an activation. Top-$k$ routing is then performed by choosing the modules according to this routing distribution.}
\vspace{-1em}
\label{fig:phatgoose}
\end{figure*}

In the meantime, extremely large-scale language models (LLMs) are now being treated as ``general-purpose'' AI systems that can perform any task without any task-specific training or adaptation.
This approach stems from the observation that LLMs often exhibit strong \textit{zero-shot generalization}, i.e.\ the ability to perform new tasks that they were not explicitly trained on.
Such zero-shot generalization capabilities are often improved through large-scale multitask fine-tuning (also called ``instruction tuning'') \citep{sanh2021multitask,wei2021finetuned,mishra2022cross}.
Relying on zero-shot generalization stands in stark contrast to the aforementioned approach of training specialized models for each task (via PEFT or otherwise).

The allure of general-purpose language models and the proliferation of specialized PEFT-based models raises a natural question:
Can we leverage a large collection of specialized modules to improve zero-shot generalization of a base language model?
Such an approach is attractive for various reasons:
First, it would provide a path to \textit{decentralized development} of generalist language models, which otherwise require a huge amount of centralized compute \citep{kaplan2020scaling,hoffmann2022training}.
In addition, it would provide a way to recycle the widespread effort and compute already being expended to create specialized models.
We might hope such an approach might be successful given the extensive evidence that multitask training improves zero-shot generalization \citep{sanh2021multitask,wei2021finetuned,mishra2022cross}, and combining specialized models could be seen as a form of multitask learning that does not require simultaneous data access.

To tackle this problem, most past work \citep{jang2023exploring,belofsky2023token,durbin2024airoboros,maxine2023llama} learns a post-hoc routing strategy by comparing an embedding of the input query to the average embedding of examples in each dataset used to train each expert.
Such methods implicitly assume there is a single expert well-suited for the query and hope that the retrieval algorithm can accurately identify this best expert.
However, \citet{jang2023exploring} showed that such approaches lag behind an ``oracle'' router that always chooses the best expert for a given query.
To explore alternative routing approaches, we first note that many PEFT methods typically insert small trainable modules in many places across the model (e.g.\ at each weight matrix \citep{hu2021lora}).
Meanwhile, many sparsely gated Mixture-of-Experts models make routing decisions separately for each token \citep{shazeer2016outrageously,fedus2022switch,du2022glam}.
In this work, we therefore explore the angle of improving zero-shot generalization through adaptive \textit{per-token} and \textit{per-module} routing. 
In doing so, the aggregate model might be better able to generalize to new tasks by using different expert capabilities at different stages and/or for different tokens.
In addition, zero-shot performance would not restrained by that of the single best specialized model and the ability to correctly retrieve it.

Building on this reasoning, we introduce \textbf{P}ost-Hoc \textbf{A}daptive \textbf{T}okenwise \textbf{G}ating \textbf{O}ver an \textbf{O}cean of \textbf{S}pecialized \textbf{E}xperts (\method), a post-hoc method that enables zero-shot generalization among specialized models. 
\method recycles PEFT modules by introducing an additional computationally inexpensive step after training the PEFT-based model itself.
Specifically, the entire model (including the newly introduced PEFT modules) is frozen and a per-module gate is trained.
This gate (whose parameters are shared across sequence positions) comprises a linear layer followed by a sigmoid nonlinearity that determines whether the activation at a given sequence position should be fed into the module or not.
Training this gate only requires a small amount of additional compute compared to performing PEFT.
The gates for every module across specialized models are then combined to determine how to route different tokens to different modules during inference using a standard ``top-$k$'' routing strategy \cite{shazeer2016outrageously,lepikhin2020gshard,du2022glam}.

To test the effectiveness of \method, we adapted T5-family models to improve zero-shot generalization on standard benchmarks.
Notably, we found that \method not only outperforms prior methods involving merging experts or retrieving a single expert but can also outperform explicit multitask training in some cases.
In qualitative analysis, we find that \method uses a diverse set of modules to perform a given task, thereby combining abilities from multiple specialized models and, in some cases, producing better performance than the single best-performing expert model.
Overall, our work sets the groundwork for a promising new framework for the decentralized development of generalist AI systems.

\section{Decentralized development of zero-shot models}
\label{sec:problem_setting}

Our goal in this work is to enable individual contributors to collectively improve zero-shot generalization capabilities of a model by sharing specialized PEFT modules.
Specifically, we formulate our problem setting as follows:
\begin{enumerate}
    \item We assume that individual contributors take a base model and perform PEFT on their specific task of interest. Since PEFT typically has lower computational and communication costs than full-model finetuning, the use of PEFT makes it easier to participate and contribute. 
    \item We assume a PEFT method introduces modules throughout the model -- for example, as discussed further in \cref{sec:phatgoose}, LoRA \citep{hu2021lora} introduces a low-rank update at every linear layer in the model. We refer to each of these updates as a ``module''.
    \item We want to avoid placing constraints on contributors or asking them to perform a large amount of additional work beyond training their PEFT-based model.
    \item Following standard practice, contributors only share their trained parameters (e.g., PEFT modules), not the dataset used for fine-tuning. Consequently, we don't allow simultaneous access to the datasets at any time, and all training on a particular dataset must be done by a single contributor.
    \item We aim to use the collection of PEFT modules to improve zero-shot performance on unseen tasks, i.e., on tasks that neither have a specialized PEFT-based model nor a training dataset. This reflects the current evaluation standard and dominant use case of LLMs.
    \item We don't aim to improve performance on held-in tasks (i.e.\ those that we have a specialized PEFT-based model for) since we can always retain performance on a given held-in task by using the expert that is specialized to that task.
\end{enumerate}

This problem setting poses many challenges.
First, while the experts are trained independently, we must determine a way to make them function together to improve performance on unseen tasks.
Second, we aim to expend as little additional compute as possible, which precludes methods that involve substantial training after the expert modules are created.
Finally, in zero-shot evaluation, the model needs to determine routing solely from information in a single input example.
This last requirement differs from past works that assume access to a target-task training dataset in order to transfer knowledge from a collection of specialized models \citep{huang2023lorahub, wu2023pi, pfeiffer2020adapterfusion, caccia2023multi,shnitzer2023large}, which we do not directly compare to.

To the best of our knowledge, most previously proposed methods that are applicable to our problem setting aim to choose a single specialized model based on properties of the input query.
The only exception we are aware of is the contemporaneous Arrow method of \citet{ostapenko2024towards}, which constructs a router using statistics of the expert parameters themselves.
We compare to Arrow in our experiments in \cref{sec:experiments}.
Among methods that route based on the input, one class of approaches (e.g.\ \citep{jang2023exploring,belofsky2023token,durbin2024airoboros,maxine2023llama}) routes to a single expert model by comparing an embedding of the input query \citep{reimers2019sentence} to the average embedding of datapoints used to train each expert.
We consider this class of methods as a primary baseline for comparison.
An additional class of methods (e.g.\ \citep{durbin2024airoboros,liu2024llamaindex}) leverage an external general-purpose LLM (e.g.\ GPT-4) and choose which model to route to by formulating a textual query that asks which model to use (e.g.\ ``I have a French model and an English model. Which model should I use for the query: combien pèse une pomme?'').
Since querying a general-purpose model (which may be suitable for processing the query itself) incurs significant expense, and because this approach has not been rigorously defined or evaluated outside of the context of open-source projects, we do not include it as a baseline.
Finally, \citet{lu2023routing} recently proposed ``Zooter'', which routes among \textit{generalist} models by training a classifier to predict which model would produce the highest-reward generation according to an auxiliary reward model.
While Zooter's goals are related to our problem setting, the fact that it involves the centralized distillation of a suitable reward model into a classifier and its focus on generalist rather than specialist models led us to exclude it as a baseline.

\section{Post-Hoc Adaptive Tokenwise Gating Over an Ocean of Specialized Experts}
\label{sec:phatgoose}

To recap, our goal in this work is to develop a method that is applicable in our problem setting -- i.e., it recycles the PEFT modules from individual contributors to improve zero-shot generalization without requiring significant extra work or simultaneous access to the contributors' datasets.
In addition, we aim to develop a method that follows the hypothesis that learning a per-token and per-module routing strategy can achieve better zero-shot generalization than a routing strategy that picks a single expert model for all tokens.
Our proposed method, \textbf{P}ost-Hoc \textbf{A}daptive \textbf{T}okenwise \textbf{G}ating \textbf{O}ver an \textbf{O}cean of \textbf{S}pecialized \textbf{E}xperts (\method), achieves these goals by having contributors train a task-specific gate and then using the parameters of each gate to perform discrete top-$k$ routing.
This process is diagrammed in \cref{fig:phatgoose}.
We now detail the specifics of these steps.

For concreteness (and in line with our experimental setting in \cref{sec:experiments}), we consider the case where contributors perform PEFT using LoRA \citep{hu2021lora}.
We emphasize that \method is applicable to any PEFT method that introduces trainable modules throughout the model (e.g.\ (IA)$^3$ \citep{liu2022few}, Adapters \citep{houlsby2019parameter}, etc.), but we consider LoRA due to its popularity and widespread use.
LoRA modifies the output of each linear layer $Wu_t$ with base model parameters $W \in \mathbb{R}^{d \times n}$ for the $t^\text{th}$ input activation $u_t \in \mathbb{R}^{n}$ as $Wu_t + BAu_t$ where $A \in \mathbb{R}^{r \times n}$ and $B\in \mathbb{R}^{d \times r}$ are trainable parameters while $W$ remains frozen during fine-tuning.
In doing so, a ``module'' (comprising a pair of $B$ and $A$ matrices) is introduced at every linear layer of the model.

After training PEFT modules on their dataset, the contributor adds a sigmoid gate layer in front of each PEFT module and trains the gate (and the gate alone, with all other parameters fixed) for a relatively small number of steps (100 in our experiments) using the same dataset and objective that was used to train the PEFT module.
The gate, which is shared across all sequence positions, determines whether or not a given activation will use the PEFT module.
In the LoRA example, a linear layer becomes $Wu_t + BAu_t\sigma(v^\mathsf{T}u_t)$ where $v \in \mathbb{R}^{n}$ is the trainable gate vector initialized to all zeros and $W$, $B$, and $A$ are all frozen.

Once the contributors share their trained PEFT modules and gate vectors, \method builds routers out of the trained gating vectors $v_a, v_b, v_c, \ldots$ to perform top-$k$ routing during inference.
A separate router is created at each layer where modules are introduced so that \method can perform per-module routing.
Specifically, we first  standardize (i.e.\ subtract the mean and divide by the standard deviation across the dimensions) both the gating vectors (denoted by $\bar{v}_a, \bar{v}_b, \ldots$) and a given activation (denoted $\bar{u}_t$) for the purpose of routing. 
Then, we assign each module a score by computing the cosine similarity between $\bar{u_t}$ and its routing vector.
We then route $u_t$ to the modules with the $k$ highest scores and, as in \citet{shazeer2016outrageously,du2022glam,lepikhin2020gshard}, rescale their output by their softmax-normalized weights.
More explicitly, during inference \method first computes the affinity $\alpha_{t, z}$ between PEFT module $z$ and activation $u_t$ as $\bar{v}_z^\mathsf{T}\bar{u}_t$. 
Then, \method assembles $\mathcal{E}_t$, the set of the top-$k$ PEFT modules for a given activation, by computing $\mathcal{E}_t = \operatorname{\mathtt{top-k}}( \alpha_{t, a}, \alpha_{t, b}, \ldots )$.
Then, scaling weights for each module are computed by $w_t = \operatorname{\mathtt{softmax}}\!\left( \left\{\alpha_{t, z}/\sqrt{n}, z \in \mathcal{E}_t \right\} \right)$
where the scaling by $1/\sqrt{n}$ is included as the typical way of avoiding saturating the softmax when fed dot products of standardized vectors \citep{vaswani2017attention}.
Finally, the output of the linear layer for activation $u_t$ is computed as $Wu_t + \sum_{z \in \mathcal{E}_t} w_{t, z}B_zA_zu_t$.

Why would sigmoid gates, trained with the rest of the model fixed, to be useful for top-$k$ style routing during inference?
We expect that the gate vector for a given PEFT module learns to associate with characteristics of activations that are associated with the task that the PEFT module is trained on. 
Combining the gates from multiple PEFT modules trained on different tasks will then route based on how relevant the corresponding PEFT module is for a given input activation. 
We note that fixing the rest of the model during gate training prevents the rest of the model from coadapting with the gate.
To validate this approach, we consider two alternate ways of forming gating vectors: First, in \cref{sec:experiments}, we consider a baseline where the router vectors are computed as the average activation over a given dataset, and second, in \cref{sec:joint}, we consider jointly training the PEFT modules and gates in one step.
Training the gate after the PEFT module has been trained and frozen ultimately leads to better performance than either of these alternatives.

To reemphasize, \method can recycle PEFT modules from contributors without requiring the datasets on which the modules were trained and without incurring significant additional costs.
However, it is important to acknowledege that \method requires an extra training step for the gates.
Despite this requirement, we find that the gate can be trained in just 100 iterations using exactly the same dataset, objective, and hyperparameters as PEFT module training, and thus imposes minimal additional burden on each contributor.  

\section{Experiments}
\label{sec:experiments}

Having introduced \method, we now turn to experimentally validating our approach.
We focus on the widely used setting of improving zero-shot generalization in T5 models.
Our experiments consider two different expert pools and three different zero-shot benchmarks.

\subsection{Setting}

\citet{sanh2021multitask} found that explicit multitask training of T5 \citep{raffel2020exploring} on a collection of prompted datasets produces a model with strong zero-shot performance on unseen tasks.
This has become a common experimental setting for benchmarking zero-shot generalization (e.g.\ \citep{chung2022scaling,longpre2023flan,jang2023exploring,zhou2022not}, etc.), so we adopt it in our study.
Specifically, as a base model, we use LM-adapted T5.1.1 XL \cite{lester2021power}, a $3$B-parameter variant of the T5 language model \citep{raffel2020exploring} that underwent an additional $100$K steps of training using a standard language modeling objective on the C4 dataset.

For creating pools of expert modules to route among, we consider two dataset collections.
For the first (``T0 Held-In''), we use the same set of 36 held-in prompted datasets and tasks that was used to train T0 \citep{sanh2021multitask}.
For the second (``FLAN''), we consider the large FLAN collection of prompted datasets \citep{longpre2023flan}.
The FLAN Collection extends T0 Held-in with zero-shot and few-shot prompted datasets from SuperGLUE \cite{wang2019superglue}, Super Natural Instructions \cite{wang2022super}, dialog datasets, and Chain-of-Thought \citep{wei2022chain} datasets.
We consider only those datasets with a zero-shot prompting format, leading to a total of $166$ specialized models from the FLAN Collection.

For our PEFT modules, we focus on Low-Rank Adapters (LoRAs, \citealp{hu2021lora}), but note that \textit{nothing about \method requires the use of LoRA} and we expect it would be equally applicable to other module architectures (e.g.\ (IA)$^3$ \citep{liu2022few}, Adapters \citep{houlsby2019parameter}, etc.).
We train a single PEFT module for each dataset in either dataset collection, leading to two settings with $36$ or $166$ expert models for T0 Held-In and FLAN respectively.

We consider three zero-shot generalization benchmarks for evaluation.
For the first (``T0HO''), we use the same held-out datasets used to evaluate T0 \citep{sanh2021multitask}.
Since the FLAN collection includes the held-out datasets from T0, we don't evaluate on T0HO when using the FLAN expert pool.
For the second and third, we consider two variants of BIG-bench \citep{srivastava2023beyond}, a community-curated collection of datasets that measure different capabilities of a model like reasoning, creativity, bias, etc.
Specifically, we evaluate on BIG-bench Hard \citep{suzgun2022challenging} and BIG-bench Lite \citep{srivastava2023beyond}. 
BIG-Bench Hard (BBH) is a collection of $23$ datasets on which state-of-the-art models performed significantly worse than humans.
BIG-Bench Lite (BBL) comprises $24$ diverse datasets that are meant as a lightweight proxy for the full BIG-Bench benchmark. 
Since the T5 tokenizer cannot tokenize some datasets in the BIG-bench collection, we exclude them during evaluation (discussion is provided in \ref{sec:big_bench}).
In all cases, we source our datasets from the Hugging Face Hub.\footnote{\url{https://huggingface.co/datasets/bigscience/P3}}\footnote{\url{https://huggingface.co/datasets/tasksource/bigbench}}\footnote{\url{https://huggingface.co/datasets/lukaemon/bbh}}

Although \method doesn't require that different contributors use the same hyperparameters, for simplicity we trained rank $r = 16$ LoRAs on every dataset for $1000$ steps on batches with $1024$ max-length-$512$ sequences using the AdamW \citep{Loshchilov2017DecoupledWD} optimizer with learning rate $5e^{-3}$ and warmup ratio of $0.06$.
We perform checkpoint selection on the validation step at a granularity of $100$ steps.
For \method, after training each module, we freeze all parameters and train the gating vector for additional $100$ steps with the same hyperparameters.
Following standard practice in past work \citep{shazeer2016outrageously,du2022glam,lepikhin2020gshard}, we use $k = 2$ for top-$k$ routing.

\subsection{Baselines}
\label{sec:baselines}

We compare against various baselines that similarly recycle expert modules in order to improve zero-shot generalization.

\paragraph{Retrieval} 
Multiple past works have considered retrieving an expert model for a given query by comparing the query's embedding \citep{reimers2019sentence} with the embeddings of examples used to train each expert \citep{jang2023exploring,maxine2023llama,durbin2024airoboros,belofsky2023token}.
Past works have differed slightly, but we found most implementation details to be relatively unimportant and based our implementation on \citet{jang2023exploring}.
Specifically, we embed text using the MiniLM-L6-v2 model\footnote{\url{https://huggingface.co/sentence-transformers/all-MiniLM-L6-v2}} \citep{reimers2019sentence}, as used by \citet{jang2023exploring,maxine2023llama,belofsky2023token}.
We store the embeddings for $1000$ random examples from each dataset used to train each expert model.
We then route each query to the expert corresponding to the example whose embedding has the highest cosine similarity to the query's embedding.

\paragraph{Average Activation} 
Since the Retrieval baseline performs example- and model-level routing (rather than token- and module-level like \method), we designed an additional baseline to compare more directly with \method.
Specifically, we consider a variant of \method where we replace each learned gating vector (e.g.\ $v_a$) with the average activation (i.e.\ the average of $u_1, u_2, \ldots$) over the dataset used to train a given expert module.
To compute the average, we use the same $1000$ random examples from each dataset as used by the Retrieval method.

\paragraph{Arrow}
\citet{ostapenko2024towards} routes among expert modules by constructing gating vectors from the expert modules themselves. 
Specifically, the method assumes the modules are LoRA experts and uses the first right singular vector of the outer product of the LoRA update $BA$ as the gating vector. 
Each input is then routed based on the probability distribution computed using the scores given by the absolute dot product between the input's representation and the gating vectors.
We use top-$k$ routing with $k=2$ in this method to have same inference compute as \method. 

\paragraph{Merged Experts}
Merging \citep{matena2022merging,choshen2022fusing}, which involves averaging the parameters of different models or modules to create a single aggregate model, provides another way to recycle models.
Recently, \citet{ostapenko2023case} demonstrated that computing a uniform average of PEFT modules can improve zero-shot generalization.
Such an approach can be seen as an extreme form of distributed or federated multitask learning \citep{smith2017federated} with a single step of federated averaging \citep{mcmahan2017communication} at the end of training.
We include merging as a baseline by computing a simple unweighted average of all of the LoRA experts in the pool.
Following \citet{ostapenko2023case}, we average after computing each LoRA outer product so that the merged modules can have a rank higher than $r$.
Note that merging is only applicable when the individual expert modules have the same architecture, whereas \method could in principle be used with heterogeneous expert modules.

Beyond the above baseline methods that satisfy our problem setting (\cref{sec:problem_setting}), we compare against a few additional baselines that violate our problem setting but nevertheless provide a useful point of comparison.

\paragraph{Multitask}
Explicit multitask training requires simultaneous access to each expert's dataset and therefore violates our problem setting.
Nevertheless, given that multitask training is a widespread and performant way to improve zero-shot generalization \citep{sanh2021multitask,wei2021finetuned}, we include it as a baseline.
We lack the computational resources to train our own multitask models, so we used publicly available models instead.
For the T0 Held-In datasets pool, we compare the T0-3B model which was trained on the same collection of datasets \citep{sanh2021multitask}.
For FLAN, there unfortunately is no public model trained on the same datasets we consider. 
The model trained on the most similar dataset mixture is FLAN-T5 XL, which includes a different (and non-public) collection of datasets.
We report the performance of FLAN-T5 XL of reference but emphasize that it should not be compared to directly.

\paragraph{Oracle}
As considered in \citep{jang2023exploring}, we consider an ``oracle'' routing scheme that chooses the specialized expert from the pool with the highest performance on a given evaluation dataset. 
Such a routing scheme is not zero-shot and serves as an upper bound on the performance of retrieval-style approaches.

\paragraph{Best Individual}
As also considered in \citep{jang2023exploring}, we find the single expert with the highest average performance across all evaluation datasets.
This approach, which is also not zero-shot, serves as the best-case performance of a degenerate routing scheme that always chooses the same expert for all inputs.

\begin{table}[t]
\centering
\footnotesize
{\setlength{\tabcolsep}{0.5em}
\begin{tabular}{@{}lcccccc@{}}
\toprule
& \multicolumn{3}{c}{\textbf{T0 Held-In}} & \multicolumn{2}{c}{\textbf{FLAN}} \\
\textbf{Method} & \textbf{T0HO} & \textbf{BBH} & \textbf{BBL} & \textbf{BBH} & \textbf{BBL} \\
\midrule
Multitask & 51.6 & 34.9 & 36.6 & \textcolor{light-gray}{38.9} & \textcolor{light-gray}{45.4} \\
Oracle & 57.2 & 42.2 & 43.5 & 45.5 & 46.5 \\
Best Individual  & 52.8 & 32.3 & 39.9 & 34.6 & 38.6 \\
\midrule
Retrieval & 50 & 30.9 & 33.6 & 31.4 & 33.1 \\
Arrow & 55.1 & 33.6 & 34.5 & 30.6 & 29.6 \\
Merged Experts & 45.4 & \textbf{35.3} & 36 & 34.6 & 34 \\
Average Activation & 50.2 & 33.8 & 35.8 & 33.5 & 34 \\
\method & \textbf{56.9} & 34.9 & \textbf{37.3} & \textbf{35.6} & \textbf{35.2} \\
\bottomrule
\end{tabular}}
\caption{Comparison of zero-shot generalization among methods built on top of LM-adapted T5.1.1 XL in two settings: T0 Held-In, where 36 experts are trained on the same datasets used to train T0 from PromptSource and FLAN, where 166 experts are trained on datasets from the FLAN collection. For T0 Held-In, we evaluate on the same held-out datasets used to evaluate T0 (T0HO) as well as BIG-Bench Hard (BBH) and Big-Bench Lite (BBL).
For a multitask baseline, we consider T0-3B for the ``T0 Held-In'' setting and FLAN-T5-XL for ``FLAN''.
Since FLAN-T5-XL was trained on a different set of datasets than the rest of the methods, we \textcolor{light-gray}{grey} out its results and caution against direct comparison.
\method generally performs best among methods that satisfy our problem setting (\cref{sec:problem_setting}) and can match or exceed the performance of explicit multitask training or non-zero-shot baselines (``Oracle'' and ``Best Individual'').
Full results on each evaluation dataset are provided in \cref{sec:T0_heldin_results} and \cref{sec:flan_results}.}
\label{tab:final_results_table}
\end{table}

\subsection{Results}

The performance of \method and the baselines we describe above on all expert pool/zero-shot benchmark combinations is presented in \cref{tab:final_results_table}.

In the T0 Held-In setting, \method generally significantly surpasses prior methods.
The improvement is especially large on T0 Held-Out tasks, where \method almost matches the performance of non-zero-shot Oracle routing.
Notably, we find that \method consistently matches or outperforms the Multitask baseline T0-3B, despite being trained in a decentralized manner (i.e.\ without simultaneous data access).
\method has slightly lower performance (0.4\%) compared to Merged Experts in BBH but outperforms it by over 11\% on T0HO and by 1.5\% on BBL.

When expanding the expert pool from 36 experts in T0 Held-in to 166 experts in the FLAN setting, \method outperforms all other routing methods on both BBL and BBH.
However, the gap between routing methods and oracle routing is generally larger in the FLAN setting.
In addition, the performance of all routing methods decreases on BBL when scaling the expert pool.
To better understand this behavior, we note that \method's performance on tasks that require logical reasoning (e.g.\ Object Counting, Ruin Names, Track Shuffled Objects, Operators, and Winowhy) tended to increase as the expert pool was scaled up, while performance on knowledge-heavy tasks (Conlang Translation, Known Unknown, Hindu Knowledge, and Novel Concepts) tended to decrease.
This could be because knowledge-intensive tasks require experts that have memorized certain information (which may be less common in a larger expert pool) whereas reasoning-based tasks benefit from composing more skills from more experts.
In any case, this behavior highlights the importance of future work on learning post-hoc routing among specialized experts for zero-shot generalization.

Beyond the specific insights covered above, we emphasize that \method consistently and significantly outperforms Retrieval and Arrow (the only prior works on post-hoc routing for zero-shot generalization).
The gate-training step in \method is essential as methods like Arrow that constructs gates using expert modules and Average Activation (a variant of \method that avoids the gate-training step) underperform compared to \method. 
In addition, Retrieval -- which lacks a mechanism to compose different experts -- consistently underperforms adaptive routing methods like \method and Average Activation across all evaluations. 
This suggests that the ability to compose knowledge from specialized experts may be beneficial when learning post-hoc routing for zero-shot generalization.

\begin{figure*}[htp]
\centering
\includegraphics[width=0.48\textwidth]{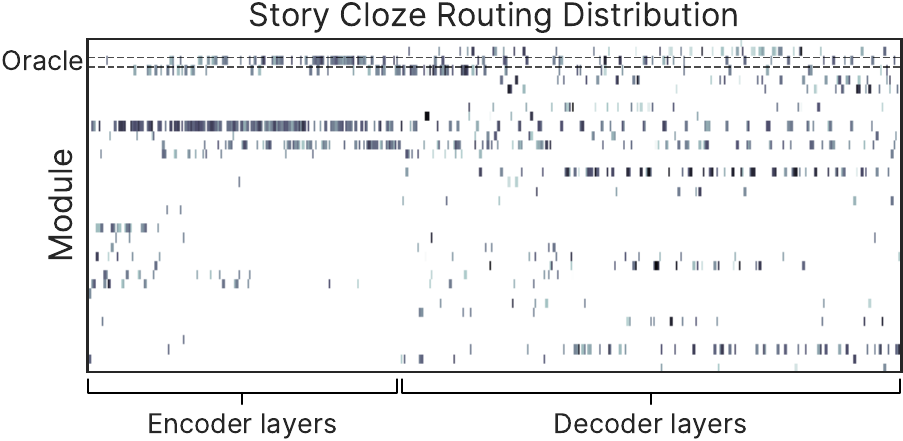}
\includegraphics[width=0.48\textwidth]{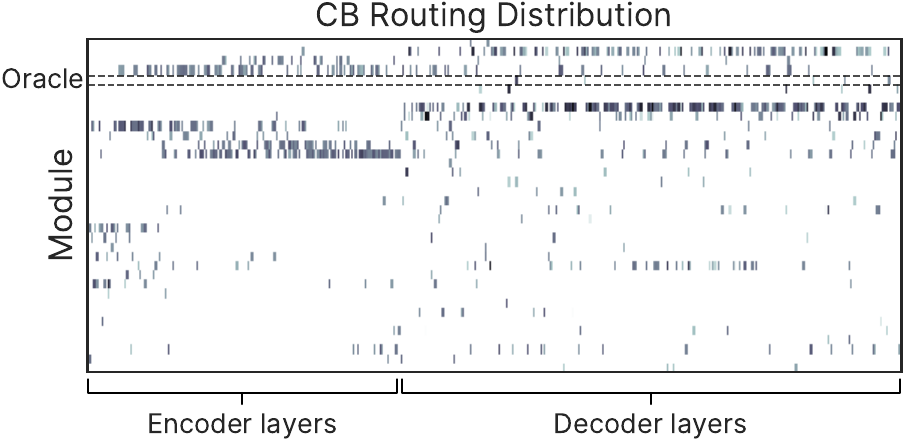}
\caption{Routing distributions produced by \method for Story Cloze and CB (from T0HO). The Oracle router's chosen module is highlighted by dashed lines. On Story Cloze, \method chooses the Oracle module in the encoder but uses diverse experts in the decoder but nevertheless matches Oracle performance. On CB, \method almost never uses the Oracle module and produces significantly better performance by using a wide range of modules.}
\label{fig:sample_routing_distributions}
\vspace{-1em}
\end{figure*}

\subsection{Qualitative Analysis}
Having established \method's strong performance, we now perform a qualitative study to better understand the benefits of adaptive per-token and per-module routing.
Specifically, we measure whether the alignment between the learned routing of \method and the Oracle routing correlates with the performance of \method. 
Given a particular evaluation dataset, we calculate the routing distribution for \method by averaging the routing probabilities of all tokens within the dataset across all modules.
We then quantify the alignment between the learned routing of \method and the Oracle using the KL divergence between their respective routing distributions. 
Then, we compute the correlation between the KL divergence of the routing distributions and the performance of \method across all datasets in our evaluation to determine whether \method's success can be attributed to alignment with Oracle routing.
We found a Pearson correlation coefficient of $-0.2$, indicating a little to no correlation between \method's performance and its alignment with Oracle.
This suggests that \method finds different performant routing strategies than Oracle routing.

To explore such strategies, in \cref{fig:sample_routing_distributions} we provide a visual representation of \method's routing distribution for two datasets, highlighting cases where \method's performance either matches or outperforms Oracle routing.
The plot illustrates the routing distribution across the set of 36 T0 Held-In modules at each layer throughout the model.
For the Story Cloze dataset, we notice that \method often routes to the same module as the Oracle in encoder layers, but uses a more diverse routing strategy in the decoder.
On CB, \method almost never routes to the Oracle module but nevertheless outperforms Oracle routing by 10\%. 
This could be attributed to \method's capability to effectively combine the capabilities of multiple experts, thereby enhancing generalization.  

\section{Related Work}

\paragraph{Routing among LLMs}
Recent work by \citet{shnitzer2023large} and \citet{lu2023routing} considers the problem of picking which generalist LLM to route a query to. 
\citet{shnitzer2023large} trains a binary classifier for each LLM to predict the correctness of its response to an input, enabling correct LLM selection during inference. 
\citet{lu2023routing} trains a router to distribute queries among LLMs, informed by distilled reward model rankings, thus avoiding the activation of all LLMs for each query.
In our work, we instead focus on routing among specialized models.

\paragraph{Recycling modules for few-shot learning}
In contrast to our focus on zero-shot generalization, some work has considered reusing specialized modules for few-shot learning on a small labeled dataset.
LoRAHub \citep{huang2023lorahub} uses a black-box optimizer to learn weights that integrate specialized LoRA modules for a few-shot task. 
In contrast, \citet{wu2024mixture} uses few-shot examples to learn a gating function with a trainable router, achieving comparable performance to LoRAHub. 
\citet{wu2023pi} trains task-specific experts, then uses task embeddings based on the diagonal of the Fisher information matrix to retrieve, average, and train modules from the top-k most similar tasks to a target task.
\citet{pfeiffer2020adapterfusion} independently learns adapters for each task, then uses a knowledge composition module to combine adapters at different layers, outperforming independent multitask training and full-model finetuning on 16 natural language understanding tasks.
\citet{gou2023mixture} trains LoRA for data clusters and a universal LoRA for the entire dataset, enhancing generalization to unseen instructions.
\citet{shah2023ziplora} trains content and style LoRAs independently, uses merge vectors to minimize interference, and combines LoRAs by using training data from both domains. 
\citet{wang2023sam} merges SAM and CLIP models to produce SAM-CLIP for language-based segmentation. 
Since all of these works use labeled target-task datasets, we exclude them from comparison.

\paragraph{Merging expert models}
Model merging \citep{choshen2022fusing,wortsman2022model,rame2022recycling,matena2022merging,ilharco2022editing,yadav2023ties,tam2023merging,jin2022dataless,yang2023adamerging} aims to combine the capabilities of models trained on different tasks or domains into a single model.
Many merging methods rely on a dataset to compute statistics or tune hyperparameters, so we focused on comparing to simple parameter averaging in our experiments (which remains a widespread method).
In addition, state-of-the-art merging methods typically underperform multitask training \citep{tam2023merging,ilharco2022editing}.
Merging has also been used as a component of systems that aim to enable zero-shot generalization.
For example, \citet{chronopoulou2023language} merges separate task and language adapters to enable cross-lingual generalization.

\paragraph{Multitask fine-tuning for zero-shot generalization}
In multitask learning, a model is trained simultaneously on a collection of datasets from different tasks.
Multitask learning generally assumes access to all datasets at once, which differs from the focus of our work.
In the simplest case, a base model is fine-tuned on a multitask mixture of datasets.
Such multitask learning has been consistently to improve zero-shot generalization on unseen tasks \citep{sanh2021multitask, chung2022scaling, wei2021finetuned}. 

\paragraph{Multitask mixture-of-expert models}
Alternatively, many recent works have explored training mixture-of-experts-style models on multitask mixtures.
In such models, a router selects the best experts for a given input, and both the router and the experts are trained using all the datasets at once. 
Studies like \citet{muqeeth2023soft, zadouri2023pushing, wang2022adamix} train a system that routes each example among a set of experts and have demonstrated improved performance on unseen tasks.
Alternatively, \citet{ponti2023combining} train a skill matrix that learns to allocate a task to a set of skills, with each skill being a parameter-efficient module. 
To adapt to a new few-shot task, they fine-tune both the skill-matrix and the experts.
\citet{caccia2023multi} show that splitting expert parameters into blocks and routing among these blocks is more effective than just routing among a set of experts.
They also find that just fine-tuning the router for a few-shot adaptation works almost as well as training both experts and the router while being more efficient. 
\citet{gupta2022sparsely} train a separate router for each task, which is a task-aware gating network. 
For a new task, they pick a router from a similar task based on domain knowledge and use it for routing examples from the new task. 
\citet{ye2022eliciting} trained a small pool of experts, each a complete transformer layer, with a router that selects different experts per layer based on task representations derived from the average embedding of dataset examples encoded using the BART encoder \cite{lewis2019bart}.
This approach allows the router to effectively select the most suitable experts for unseen tasks by leveraging task-specific representations.
These past works on multitask mixture-of-experts models bear some similarity to our problem setting but ultimately rely on simultaneous data access.
However, we are optimistic that insights could be shared between these complementary settings.


\section{Conclusion}

In this paper, we introduced \textbf{P}ost-\textbf{H}oc \textbf{A}daptive \textbf{T}okenwise \textbf{G}ating \textbf{O}ver an \textbf{O}cean of \textbf{S}pecialized \textbf{E}xperts (\method).
\method provides a way to recycle expert modules created through parameter-efficient training to improve zero-shot generalization of a base model.
Specifically, \method has contributors perform an additional computationally inexpensive step that involves training a sigmoid gate for each module.
The parameters of these gates are combined to produce a top-$k$ router among modules.
In experiments on the widely used setting of improving zero-shot generalization of T5-family models, we found that \method generally outperforms other methods that learn post-hoc routing strategies among specialized modules and frequently matches or outperforms explicit multitask training.
We also qualitatively analyzed the routing learned by \method and found that it can learn performant routing strategies that differ from a simple Oracle strategy that routes to the module that attains the best performance on a given task.

Our work, and our proposed problem setting, open up avenues for future work on decentralized collaborative model development.
First, while we focused on the standard setting of adapting T5-family models for better zero-shot generalization, we would be interested in applying \method to decoder-only Transformers that have become widespread in the development of LLMs.
Second, while our investigation centered on LoRA-based modules with the same rank \cite{hu2021lora}, \method is applicable to a wide range of module architectures, including cases where modules do not necessarily share an architecture.
Exploring different PEFT module architectures (such as Adapters \cite{houlsby2019parameter} and (IA)$^3$ \cite{liu2022few}) and routing among heterogeneous models could improve the performance and efficiency of \method. 
Finally, we note again that none of the post-hoc routing strategies we considered exhibited consistent gains when increasing the size of the module collection. 
This pattern mirrors trends noted in explicit multitask training, where models exhibit strong performance on certain datasets while underperforming on others \cite{sanh2021multitask, chung2022scaling}.
Addressing this could significantly contribute to the development of models that not only learn continually but also exhibit enhanced zero-shot generalization for unseen tasks as the expert pool expands.
Overall, we are optimistic that future work will study and build on these issues and enable a new paradigm for model development.

\section*{Impact Statement}
This paper presents work whose goal is to advance the field of Machine Learning. 
There are many potential societal consequences 
of our work, none which we feel must be specifically highlighted here.

\section*{Acknowledgements}
Thanks to Derek Tam for feedback on a draft of this paper. This work was supported by NSF-AI Engage Institute DRL-2112635.

\bibliography{icml2024/example_paper}
\bibliographystyle{icml2024/icml2024}

\newpage
\appendix
\onecolumn
\section{Joint training of gates and expert parameters}
\label{sec:joint}
While \method trains the gate as a separate step after training the expert module, we also explored the possibility of jointly training both the gate and module. 
Specifically, within the T0 Held-in setting utilizing the LM adapted T5.1.1 Large model, we found that joint training underperforms \method, exhibiting a notable 5.6\% decrease in performance on T0 evaluation. 
Full results comparing these methods on T0 Held-Out datasets are detailed in \cref{tab:T0_results_jointtrain_ablation}.
A possible explanation for this performance drop could be that model might update the module parameters in such a way that reduces the effect of scaling introduced by gates and consequently making the gates less useful when combined across modules from different datasets. 
\begin{table}[h]
\centering
\footnotesize
{\setlength{\tabcolsep}{0.5em}
\begin{tabular}{l c c c c c c c c c c c c}
\toprule
\textbf{Method} & \textbf{Avg} & \textbf{RTE} & \textbf{H-Swag} & \textbf{COPA} & \textbf{WIC} & \textbf{\makecell{Wino\\grande}} & \textbf{CB} & \textbf{\makecell{Story\\Cloze}} & \textbf{ANLI-R1} & \textbf{ANLI-R2} & \textbf{ANLI-R3} & \textbf{WSC} \\
\midrule
Joint training & 46.2 & 54.5 & 25.3 & 57.1 & 50.2 & 52.3 & 51.2 & 60.6 & 30.4 & 31.7 & 32 & 62.9 \\ 
\method & 51.8 & 62.7 & 28	& 73.1 & 50.5 & 53 & 54.9 & 90.6 & 30 & 31.3 & 32.6	& 63.4 \\
\bottomrule
\end{tabular}}
\caption{ Results on T0 Held-Out datasets. Both the methods use expert modules trained from LM-adapted T5.1.1 Large on T0 Held-in datasets. Joint training trains gate and module parameters at the same time, whereas \method trains gate parameters post-hoc freezing the trained module and pretrained backbone. A considerable drop in performance with joint training indicates that gates are effective when learned post-hoc as in \method}
\label{tab:T0_results_jointtrain_ablation}
\end{table}

\section{BIG-bench datasets}
\label{sec:big_bench}
We removed certain datasets from BIG-bench during evaluation as they cannot be tokenized by the T5 tokenizer. 
Specifically, one dataset (Dyck Languages) is removed from BIG-bench Hard benchmark and seven datasets (Auto 
Debugging, Code Line Description, Emoji Movie, Language Identification, Misconceptions Russian, Parsinlu Reading Comprehension, and Symbol Interpretation) are removed from the BIG-bench Lite benchmark. 
Most of these datasets either have emojis or curly braces or are non-English languages, none of which can be tokenized using the T5 tokenizer. 

\section{Results from T0 Held-in setting}
\label{sec:T0_heldin_results}
\begin{table}
\centering
\begin{adjustbox}{angle=90}
\begin{tabular}{l c c c c c c c c c c c c}
\toprule
\textbf{Method} & \textbf{Avg} & \textbf{RTE} & \textbf{H-Swag} & \textbf{COPA} & \textbf{WIC} & \textbf{Winogrande} & \textbf{CB} & \textbf{StoryCloze} & \textbf{ANLI-R1} & \textbf{ANLI-R2} & \textbf{ANLI-R3} & \textbf{WSC} \\
\midrule
T0 3B & 51.6 & 60.1 & 26.9 & 74.8 & 51.3 & 50.9 & 52.7 & 85.1 & 34.7 & 33 & 33.5 & 64.9 \\
Oracle & 57.2 & 66.9 & 36.8 & 89.6 & 52.4 & 57.6 & 59.9 & 96.9 & 34.5 & 34.8 & 36.7 & 63.6 \\
Best Individual & 52.8 & 57.6 & 28.6 & 85.1 & 50.0 & 54.1 & 56.0 & 94.2 & 34.4 & 34.3 & 34.8 & 52.1 \\
Retrieval & 50 & 55.7 & 27.7 & 73 & 49.8 & 52.9 & 51.9 & 79.8 & 32.8 & 34.2 & 33.7 & 58.8 \\
Arrow & 55.1 & 70.8 & 28.1 & 81.2 & 52.0 & 57.0 & 73.6 & 83.0 & 35.6 & 35.0 & 37.0 & 52.4 \\
\makecell[l]{Merged Experts \\- parameter average} & 42 & 52.7 & 23.1 & 57 & 49.8 & 51.3 & 36 & 49.4 & 33.1 & 33.4 & 33.1 & 42.9 \\
Merged Experts & 45.4 & 55.7 & 25.7 & 61.8 & 50.3 & 53.3 & 45.6 & 63.8 & 33.1 & 33.4 & 33.4 & 43.5 \\
 Average activation & 50.2 & 53.2 & 25.8 & 71.3 & 50.3 & 58.5 & 61.7 & 80.1 & 33.9 & 33.9 & 35.1 & 48.9 \\
\method & 56.9 & 65.9 & 29.1 & 91.1 & 50.6 & 59.9 & 71.7 & 96.3 & 35.4 & 34 & 38.1 & 53.8 \\
\bottomrule
\end{tabular}
\end{adjustbox}
\caption{ Complete results on  T0 Held-Out datasets. }
\label{tab:T0_results_all_methods}
\end{table}

\begin{sidewaystable}
\small
\centering
\begin{tabular}{l c c c c c c c c}
\toprule
\textbf{Expert} & \textbf{Avg} & \textbf{\makecell{Boolean\\Expression}} & \textbf{\makecell{Causal\\Judgment}} & \textbf{\makecell{Date \\Understanding}} & \textbf{\makecell{Disambiguator\\QA}} & \textbf{\makecell{Formal\\Fallacies}} & \textbf{\makecell{Geometric\\Shapes}} & \textbf{Hyperbaton} \\
\midrule
Multitask & 34.9 & 49.6 & 55.3 & 35.2 & 55.4 & 51.5 & 10.6 & 50 \\
Oracle & 42.2 & 64.4 & 59.5 & 41.7 & 65.1 & 51.7 & 30.1 & 52.7 \\
Best Individual & 32.3 & 47.6 & 49.5 & 36.9 & 57 & 51.7 & 20.1 & 49.9 \\
Retrieval & 30.9 & 44.8 & 46.3 & 26.8 & 48.4 & 50.7 & 18.4 & 49.5 \\
Arrow & 33.6 & 57.2 & 56.8 & 36.0 & 46.9 & 49.6 & 15.3 & 50.0 \\
\makecell[l]{Merged Experts \\- parameter average} & 31.9 & 48.8 & 53.2 & 35.8 & 43 & 50 & 19.8 & 51.2 \\
Merged Experts & 35.3 & 60.8 & 58.4 & 38.5 & 45.3 & 50.1 & 10 & 50 \\
Average Activation & 33.8 & 57.2 & 56.8 & 37.4 & 41.5 & 50 & 10.3 & 49.1 \\
\method & 34.9 & 52 & 57.4 & 39 & 58.1 & 50.1 & 9.5 & 50 \\
\end{tabular}

\begin{tabular}{l c c c c c c c c}
\toprule
\textbf{Expert} & \textbf{\makecell{Logical\\Detection}} & \textbf{\makecell{Movie\\Recommendation}} & \textbf{\makecell{Multistep\\Arithmetic}} & \textbf{Navigate} & \textbf{\makecell{Object\\Counting}} & \textbf{\makecell{Penguins\\in a\\Table}} & \textbf{\makecell{Reasoning \\about Colored\\Objects}} & \textbf{\makecell{Ruin\\Names}} \\
\midrule
Multitask & 47.9 & 34.8 & 0 & 50 & 22.5 & 32.9 & 42 & 19.6 \\
Oracle & 45.8 & 49.2 & 1.6 & 50 & 28.1 & 36.9 & 53.2 & 49.6 \\
Best Individual & 45.8 & 23.8 & 0.4 & 50 & 0 & 34.2 & 43.4 & 15.2 \\
Retrieval & 33.3 & 34.4 & 0.4 & 50 & 2.4 & 15.4 & 36.1 & 20.8 \\
Arrrow & 39.7 & 42.4 & 0.8 & 50 & 0 & 33.6 & 47.1 & 13.8 \\
\makecell[l]{Merged Experts \\- parameter average} & 39.8 & 26.2 & 0 & 50 & 0.6 & 28.2 & 27.6 & 25 \\
Merged Experts & 44.3 & 23 & 0.4 & 50 & 25.4 & 35.6 & 47.5 & 26.6 \\
Average Activation & 27.7 & 23.4 & 0.8 & 50 & 4.5 & 34.2 & 46.8 & 35 \\
\method & 40.5 & 30.8 & 0.8 & 50 & 20.4 & 30.9 & 47.8 & 26.8 \\
\end{tabular}

\begin{tabular}{l c c c c c c c}
\toprule
\textbf{Expert} & \textbf{\makecell{Salient\\Translation\\Error\\Detection}} & \textbf{Snarks} & \textbf{\makecell{Sports\\Understanding}} & \textbf{\makecell{Temporal\\Sequences}} & \textbf{\makecell{Track \\Shuffled\\Objects}} & \textbf{\makecell{Web of\\Lies}} & \textbf{\makecell{Word\\Sorting}}\\
\midrule
Multitask & 27.8 & 46.4 & 50.2 & 16.1 & 17.4 & 51.6 & 0.5 \\
Oracle & 27 & 61.3 & 50.9 & 28.2 & 20.1 & 59.2 & 2.8 \\
Best Individual & 15.4 & 37.6 & 50.3 & 12.7 & 16 & 51.2 & 1.1 \\
Retrieval & 16.5 & 51.9 & 50.3 & 18.7 & 19.3 & 44.8 & 0.1 \\
Arrow & 25.1 & 45.3 & 49.9 & 13.1 & 17.1 & 50.4 & 0 \\
\makecell[l]{Merged Experts \\- parameter average} & 13.5 & 43.6 & 50.3 & 26.3 & 16.5 & 48.8 & 2.8 \\
Merged Experts & 24.9 & 44.8 & 51.7 & 19.5 & 17 & 51.6 & 0.9 \\
Average Activation & 28.1 & 46.4 & 50.1 & 23.7 & 17.3 & 52 & 1.3 \\
\method & 25.5 & 45.3 & 51.3 & 12.3 & 16.2 & 53.2 & 0 \\
\bottomrule
\end{tabular}

\caption{BIG-bench Hard (BBH) results of different methods in T0 Held-In setting}
\label{tab:bbh_results_methods_t0}
\end{sidewaystable}

\begin{sidewaystable}
\small
\centering
\begin{tabular}{l c c c c c c}
\toprule
\textbf{Expert} & \textbf{Avg} & \textbf{\makecell{BBQ Lite\\Json}} & \textbf{\makecell{Conceptual\\Combinations}} & \textbf{\makecell{Conlong\\Translation}} & \textbf{\makecell{Formal\\Fallacies}} & \textbf{\makecell{Hindu\\Knowledge}} \\
\midrule
Multitask & 36.6 & 40.8 & 44.7 & 26 & 51.5 & 40.6 \\
Oracle & 43.5 & 55.3 & 62.1 & 29.8 & 51.6 & 46.3 \\
Best Individual & 39.9 & 54.5 & 47.6 & 29.6 & 51.6 & 44.6 \\
Retrieval & 33.6 & 44 & 31.1 & 7.9 & 50.7 & 36.6 \\
Arrow & 34.5 & 45.5 & 32 & 8.9 & 49.6 & 37.7 \\
\makecell[l]{Merged Experts \\- parameter average} & 32.8 & 38.2 & 24.3 & 26.5 & 50 & 28.6 \\
Merged Experts & 36 & 42.5 & 33 & 28.9 & 50.1 & 40 \\
Average Activation & 35.8 & 43.3 & 27.2 & 26.6 & 50 & 41.1 \\
\method & 37.3 & 48 & 38.8 & 25.1 & 50 & 47.4 \\
\end{tabular}

\begin{tabular}{l c c c c c c}
\toprule
\textbf{Expert} & \textbf{\makecell{Known\\Unknowns}} & \textbf{\makecell{Linguistic\\Puzzles}} & \textbf{\makecell{Logic\\Grid\\Puzzle}} & \textbf{\makecell{Logical\\Detection}} & \textbf{\makecell{Novel\\Concepts}} & \textbf{Operators} \\
\midrule
Multitask & 47.8 & 0 & 35.9 & 48.1 & 40.6 & 1 \\
Oracle & 65.2 & 0 & 41.7 & 45.4 & 43.8 & 8.6 \\
Best Individual & 60.9 & 0 & 39.9 & 45.4 & 34.4 & 3.3 \\
Retrieval & 58.7 & 0 & 33.7 & 33.3 & 34.4 & 1 \\
Arrow & 50	& 0	& 42.6	& 39.7	& 31.2	& 1 \\
\makecell[l]{Merged Experts \\- parameter average} & 50 & 0 & 37 & 39.8 & 40.6 & 2.9 \\
Merged Experts & 45.7 & 0 & 39.6 & 44.3 & 28.1 & 7.1 \\
Average Activation & 45.7 & 0 & 34.9 & 27.7 & 37.5 & 5.2 \\
\method & 52.2 & 0 & 39.3 & 40.5 & 34.4 & 2.4 \\
\end{tabular}

\begin{tabular}{l c c c c c c}
\toprule
\textbf{Expert} & \textbf{\makecell{Play Dialog\\Same or \\Different}} & \textbf{\makecell{Repeat Copy\\Logic}} & \textbf{\makecell{Strange\\Stories}} & \textbf{\makecell{Strategy \\QA}} & \textbf{\makecell{Vitamin C Fact\\Verification}} & \textbf{Winowhy} \\
\midrule
Multitask & 45.8 & 0 & 47.7 & 52.5 & 54.2 & 44.3 \\
Oracle & 63.3 & 0 & 68.4 & 56.1 & 51.1 & 50.5 \\
Best Individual & 63.1 & 0 & 60.3 & 53.6 & 44.6 & 44.2 \\
Retrieval & 42.9 & 0 & 54.6 & 52.9 & 44 & 44.7 \\
Arrow & 36.9	& 0	& 52.9	& 52.2	& 62.4	& 44.3 \\
\makecell[l]{Merged Experts \\- parameter average} & 37.8 & 0 & 43.1 & 52.8 & 41 & 44.3 \\
Merged Experts & 36.9 & 0 & 56.3 & 54.3 & 61.3 & 44.3 \\
Average Activation & 49.9 & 0 & 64.4 & 53.3 & 56.6 & 44.7 \\
\method & 37 & 0 & 65.5 & 52.6 & 57.2 & 44.4 \\
\bottomrule
\end{tabular}

\caption{BIG-bench Lite (BBL) results of different methods in T0 Held-In setting}
\label{tab:bblite_results_methods_t0}
\end{sidewaystable}

\section{Results from FLAN setting}
\label{sec:flan_results}
\begin{sidewaystable}
\small
\centering
\begin{tabular}{l c c c c c c c c}
\toprule
\textbf{Expert} & \textbf{Avg} & \textbf{\makecell{Boolean\\Expression}} & \textbf{\makecell{Causal\\Judgment}} & \textbf{\makecell{Date \\Understanding}} & \textbf{\makecell{Disambiguator\\QA}} & \textbf{\makecell{Formal\\Fallacies}} & \textbf{\makecell{Geometric\\Shapes}} & \textbf{Hyperbaton} \\
\midrule
Multitask & 38.9 & 50 & 61.1 & 36.6 & 65.9 & 52.2 & 9.7 & 51.1 \\
Oracle & 45.5 & 66 & 59.5 & 42.3 & 65.1 & 52.9 & 30.1 & 69.3 \\
Best Individual & 34.6 & 52.8 & 47.9 & 39 & 52.3 & 50 & 10.3 & 50.2 \\
Retrieval & 31.4 & 50.4 & 45.8 & 31.7 & 39.9 & 50.3 & 18.4 & 48.9 \\
Arrow & 30.6& 54.4 & 52.6 & 24.1 & 30.2	& 50.1	& 8.9	& 51.7 \\
\makecell[l]{Merged Experts \\- parameter average} & 34.3 & 52.4 & 53.7 & 35.8 & 40.7 & 50.9 & 20.1 & 51.1 \\
Merged Experts & 34.6 & 53.6 & 56.8 & 36.9 & 45.7 & 50 & 12 & 52.2 \\
Average Activation & 33.5 & 56 & 54.2 & 33.6 & 31.8 & 49.2 & 10.9 & 50.2 \\
\method & 35.6 & 51.6 & 57.9 & 34.1 & 57.4 & 50.5 & 10.3 & 48.5 \\
\end{tabular}

\begin{tabular}{l c c c c c c c c}
\toprule
\textbf{Expert} & \textbf{\makecell{Logical\\Detection}} & \textbf{\makecell{Movie\\Recommendation}} & \textbf{\makecell{Multistep\\Arithmetic}} & \textbf{Navigate} & \textbf{\makecell{Object\\Counting}} & \textbf{\makecell{Penguins\\in a\\Table}} & \textbf{\makecell{Reasoning \\about Colored\\Objects}} & \textbf{\makecell{Ruin\\Names}} \\
\midrule
Multitask & 49.6 & 32.8 & 0 & 50 & 35.7 & 39.6 & 56.6 & 19 \\
Oracle & 48.8 & 49.2 & 1.6 & 54.6 & 45.7 & 37.6 & 53.5 & 49.6 \\
Best Individual & 41.3 & 22.8 & 0.4 & 50 & 16.4 & 29.5 & 36.5 & 35.3 \\
Retrieval & 28.5 & 35 & 0 & 50 & 16 & 24.8 & 21.1 & 23.4 \\
Arrow & 24	& 30.4	& 1.2	& 50	& 0& 	28.9 &	43.1& 	15.2 \\
\makecell[l]{Merged Experts \\- parameter average} & 42.8 & 23 & 0.4 & 50 & 24.6 & 30.9 & 38.4 & 25.2 \\
Merged Experts & 42.9 & 23.2 & 0.8 & 50 & 24.1 & 34.2 & 44.5 & 28.3 \\
Average Activation & 26.9 & 26.2 & 0.4 & 50 & 20.1 & 38.3 & 41.7 & 23 \\
\method & 41.4 & 26.8 & 0.8 & 50 & 41 & 32.9 & 47.2 & 34.4 \\
\end{tabular}

\begin{tabular}{l c c c c c c c}
\toprule
\textbf{Expert} & \textbf{\makecell{Salient\\Translation\\Error\\Detection}} & \textbf{Snarks} & \textbf{\makecell{Sports\\Understanding}} & \textbf{\makecell{Temporal\\Sequences}} & \textbf{\makecell{Track \\Shuffled\\Objects}} & \textbf{\makecell{Web of\\Lies}} & \textbf{\makecell{Word\\Sorting}}\\
\midrule
Multitask & 39.2 & 59.7 & 51.2 & 27.2 & 15.7 & 53.6 & 0 \\
Oracle & 31.5 & 61.3 & 52.5 & 45.3 & 20.3 & 61.6 & 2.9 \\
Best Individual & 22.3 & 55.8 & 50.3 & 27.1 & 15.5 & 53.6 & 1.9 \\
Retrieval & 17.7 & 51.4 & 51.8 & 18.6 & 18.3 & 47.6 & 0.3 \\
Arrow & 19.9	& 51.9	& 50.7	& 16.4	& 18.9	& 51.6	& 0 \\
\makecell[l]{Merged Experts \\- parameter average} & 27.2 & 44.2 & 50.4 & 26 & 16.2 & 50 & 0.9 \\
Merged Experts & 26.4 & 40.9 & 49.9 & 22.4 & 16.3 & 49.6 & 1.2 \\
Average Activation & 26.6 & 47.5 & 50.7 & 26.7 & 18.1 & 53.2 & 1.2 \\
\method & 21 & 47 & 49.9 & 9.3 & 17 & 54 & 0.2 \\
\bottomrule
\end{tabular}

\caption{BIG-bench Hard (BBH) results of different methods in the FLAN setting.}
\label{tab:bbh_results_methods_flan}
\end{sidewaystable}

\begin{sidewaystable}
\small
\centering
\begin{tabular}{l c c c c c c}
\toprule
\textbf{Expert} & \textbf{Avg} & \textbf{\makecell{BBQ Lite\\Json}} & \textbf{\makecell{Conceptual\\Combinations}} & \textbf{\makecell{Conlong\\Translation}} & \textbf{\makecell{Formal\\Fallacies}} & \textbf{\makecell{Hindu\\Knowledge}} \\
\midrule
Multitask & 45.4 & 66.9 & 72.8 & 27.9 & 52.2 & 40 \\
Oracle & 46.5 & 61.3 & 62.1 & 36 & 52.9 & 46.3 \\
Best Individual & 38.6 & 61.3 & 50.5 & 28.3 & 50 & 41.7 \\
Retrieval & 33.1 & 35.8 & 35.9 & 10.8 & 50.3 & 25.1 \\
Arrrow & 29.6	& 35.3 &	19.4	& 0.2	& 50.1	& 34.3 \\
\makecell[l]{Merged Experts \\- parameter average} & 33.7 & 39.2 & 26.2 & 31 & 50.9 & 31.4 \\
Merged Experts & 34 & 40.2 & 27.2 & 29.1 & 50 & 36.6 \\
Average Activation & 34 & 41 & 28.2 & 6.6 & 49.2 & 38.3 \\
\method & 35.2 & 49 & 37.9 & 13.2 & 50.5 & 42.3 \\
\end{tabular}

\begin{tabular}{l c c c c c c}
\toprule
\textbf{Expert} & \textbf{\makecell{Known\\Unknowns}} & \textbf{\makecell{Linguistic\\Puzzles}} & \textbf{\makecell{Logic\\Grid\\Puzzle}} & \textbf{\makecell{Logical\\Detection}} & \textbf{\makecell{Novel\\Concepts}} & \textbf{Operators} \\
\midrule
Multitask & 58.7 & 0 & 42.8 & 49.9 & 37.5 & 13.3 \\
Oracle & 65.2 & 0 & 42.5 & 48.6 & 50 & 12.4 \\
Best Individual & 56.5 & 0 & 40.8 & 41.3 & 28.1 & 6.7 \\
Retrieval & 47.8 & 0 & 34.2 & 28.5 & 34.4 & 8.6 \\
Arrow & 54.3	& 0	& 31.8 & 	24	& 31.2 &	1.9 \\
\makecell[l]{Merged Experts \\- parameter average} & 50 & 0 & 35.9 & 42.8 & 37.5 & 8.1 \\
Merged Experts & 43.5 & 0 & 37.9 & 42.9 & 34.4 & 7.6 \\
Average Activation & 54.3 & 0 & 35 & 26.9 & 34.4 & 7.6 \\
\method & 37 & 0 & 35.6 & 41.4 & 25 & 5.2 \\
\end{tabular}

\begin{tabular}{l c c c c c c}
\toprule
\textbf{Expert} & \textbf{\makecell{Play Dialog\\Same or \\Different}} & \textbf{\makecell{Repeat Copy\\Logic}} & \textbf{\makecell{Strange\\Stories}} & \textbf{\makecell{Strategy \\QA}} & \textbf{\makecell{Vitamin C Fact\\Verification}} & \textbf{Winowhy} \\
\midrule
Multitask & 44.4 & 0 & 75.9 & 65.7 & 78.5 & 45.3 \\
Oracle & 63.3 & 0 & 74.1 & 56.1 & 66.7 & 53.8 \\
Best Individual & 45.9 & 0 & 74.1 & 53.9 & 33 & 44.4 \\
Retrieval & 43.2 & 0 & 55.2 & 51.3 & 50.4 & 50.7 \\
Arrow & 37.1	& 0	& 37.9	& 50.5	& 50.2	& 44.5 \\
\makecell[l]{Merged Experts \\- parameter average} & 36.9 & 0 & 43.1 & 50.1 & 44.9 & 44.3 \\
Merged Experts & 36.9 & 0 & 46.6 & 52.1 & 47.9 & 44.3 \\
Average Activation & 38.1 & 0 & 59.2 & 51.2 & 61.3 & 46.1 \\
\method & 37.3 & 0 & 63.8 & 51.2 & 62.6 & 47 \\
\bottomrule
\end{tabular}

\caption{BIG-bench Lite (BBL) results of different methods in the FLAN setting.}
\label{tab:bblite_results_methods_flan}
\end{sidewaystable}

\end{document}